\documentclass[letterpaper, 10 pt, conference]{ieeeconf}  

\IEEEoverridecommandlockouts                              

\usepackage{xcolor}
\usepackage{comment}
\usepackage{paralist}
\usepackage{hyperref}
\usepackage{graphicx}
\usepackage{amsfonts}
\usepackage{amsmath}
\usepackage[english]{babel}
\usepackage{booktabs}
\usepackage{multirow}
\usepackage[skip=2pt]{caption} 
\usepackage{subcaption}
\usepackage{floatrow}
\usepackage{sidecap}
\usepackage[affil-it]{authblk}

\title{\LARGE \bf

Reinforcement Learning for Picking Cluttered General Objects with Dense Object Descriptors

}

\author[1,2]{Hoang-Giang Cao$^*$}
\author[1]{Weihao Zeng$^*$}
\author[1,2]{I-Chen Wu$^\dagger$}
\affil[1]{Department of Computer Science, National Yang Ming Chiao Tung University, Taiwan}
\affil[2]{Research Center for IT Innovation, Academia Sinica, Taiwan}

\begin{document}

\maketitle

\def\thefootnote{*}\footnotetext{Equal contribution.}\def\thefootnote{\arabic{footnote}}
\def\thefootnote{$\dagger$}\footnotetext{Correspondence.}\def\thefootnote{\arabic{footnote}}

\thispagestyle{empty}
\pagestyle{empty}

\definecolor{TodoColor}{rgb}{1.0, 0.0, 1.0}

\newcommand{\todo}[1]{{\color{TodoColor} [TODO: #1]}}
\begin{abstract}

Picking cluttered general objects is a challenging task due to the complex geometries and various stacking configurations.
Many prior works utilize pose estimation for picking, but pose estimation is difficult on cluttered objects.
In this paper, we propose Cluttered Objects Descriptors (CODs), a  dense  cluttered  objects  descriptor which can represent rich object structures, and use the pre-trained CODs network along with its intermediate outputs to train a picking policy.
Additionally, we train the policy with reinforcement learning, which enable the policy to learn picking without supervision.
We conduct experiments to demonstrate that our CODs is able to consistently represent seen and unseen cluttered objects, which allowed for the picking policy to robustly pick cluttered general objects.
The resulting policy can pick 96.69\% of unseen objects in our experimental environment that are twice as cluttered as the training scenarios.
\end{abstract}
\section{Introduction}

Recently, intelligent robotics systems are of great interests, and manipulation skills are vital for such systems. Grasping, one of the most important manipulation skills, is essential for many real-world applications.
For example, with pick-and-place, industrial robots can offset laborious industrial routines from human workers, and household robots can help with chores.
Despite the importance of robotics grasping, it remains as an open problem \cite{kumra2017robotic,zeng2018robotic}.

With the recent developments in data-driven methods, deep learning is a promising direction for solving the grasping problem.
Most current grasping methods require dense labeling, either manually labeled or analytically generated \cite{du2021vision}.
Such methods generally generate labels for grasping, and treat the grasping problem as a supervised learning problem.
For example, Mahler used Quality Convolutional Neural Network (GQ-CNN) to learn to predict analytically generated robustness score based on geometry and physics information \cite{mahler2018dex, mahler2019learning}.
With the recent development in deep reinforcement learning (DRL), many self-supervised end-to-end methods have been proposed.
DRL methods allow for end-to-end training without supervision.
Kalashnikov {\it et al.} \cite{kalashnikov2018qt} proposed QT-Opt, a vision based robotic manipulation system on real robots using DRL.
QT-Opt demonstrated that DRL methods could automatically learn appealing behaviors, such as re-grasping and pre-grasp.
DRL is a particular promising approach for solving the grasping problem.

\begin{figure}[ht!b]
\begin{center}
  \includegraphics[width=\columnwidth]{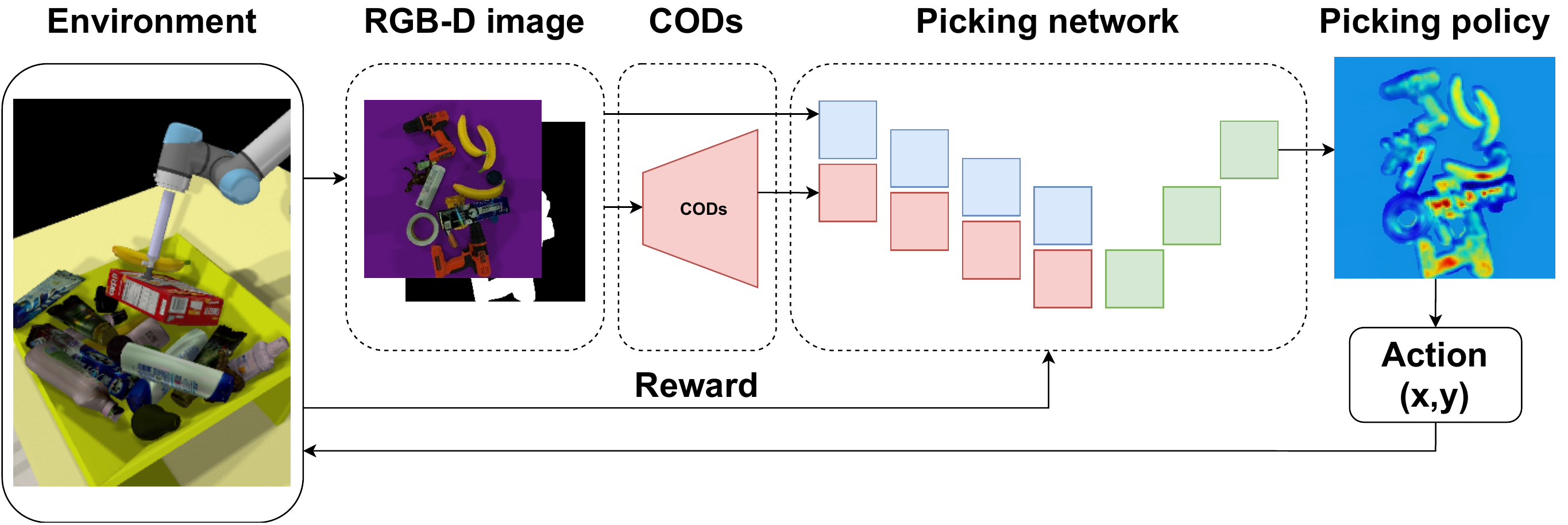}
  \caption{The training pipeline.} 
  \label{fig:training_pipeline}
\end{center}  
\end{figure}

Suction grasping, despite its reliability and simplicity, attracts far less attention than other types of grasps in the research community \cite{cao2021suctionnet, du2021vision}.
Zeng {\it et al.} manually labeled RGB-D images for training suction grasp network in the 2017 Amazon Robotics Challenge \cite{zeng2018robotic}.
Shao {\it et al.} proposed a self-supervised method for suction grasping in clutter environments, but Shao only considered cylinders \cite{shao2019suction}.
However, a better solution is still lacking to use suction grasping for picking cluttered general objects. 

A good representation is critical to achieve good performance for DRL methods \cite{du2019good}.
Florence {\it et al.} proposed Dense Object Nets (DONs), which generate dense descriptors with rich object structure information \cite{florence2018dense}.
Chai {\it et al.} used DONs to accomplish pick-and-place from a single demonstration \cite{chai2019multi}.
Ganapathi \cite{ganapathi2020learning} and Sundaresan \cite{sundaresan2020learning} used DONs to perform soft material manipulation.
To better represent cluttered general objects in the grasping problem, we utilize DONs to generate representations.

In this paper, we present a novel method capable of picking general objects from cluttered environment.
We obtain the Cluttered Objects Descriptors (CODs) network using DONs, and train a picking policy on top of it.
We employ Actor-Critic \cite{mnih2016asynchronous}, a reinforcement learning method, to train the picking policy.
Inspired by the method in \cite{shao2019suction}, we use the intermediate output of the trained CODs network together with the RGB-D input to train the picking policy.

\textbf{Contributions.} The main contributions of this paper can be summarized as follows:
\begin{inparaenum}[1)]
    \item We extend DONs \cite{florence2018dense} to CODs, a dense cluttered objects descriptor which can consistently represent cluttered objects with generalization ability.
    \item We propose a novel DRL approach that employ intermediates outputs of the trained CODs network to better pick cluttered general objects.
    \item In the experiments, we demonstrate that our method outperforms other methods, and can be generalized to grasping unseen objects with more cluttered scenarios.
\end{inparaenum}

\section{Related Works}
In this section, we will review some existing grasping methods and the Dense Object Nets.

\subsection{Grasping Methods}

Grasping is widely studied by the robotics research community.
Some previous methods generated annotated datasets and treated suction grasping as a supervised learning problem \cite{zeng2018robotic, utomo2021suction, jiang2020depth, mahler2018dex, hasegawa2019graspfusion}. Zeng {\it et al.}, the first-place winner of the Amazon Robotics Challenge, manually labeled RGB-D images \cite{zeng2018robotic}.
Qin {\it et al.} generated grasps with scores using a simulator and analytical methods \cite{qin2020s4g}.
Mahler {\it et al.} generated suction labels analytically by considering quality of vacuum seals \cite{mahler2018dex}.
Danielczuk {\it et al.} introduced an algorithm based on learned occupancy distributions for grapsing object with both parallel gripper and suction pad \cite{danielczuk2020xray}.
Some other methods calculated the best suction grasp analytically\cite{valencia20173d,wan2020planning}.

With the recent developments in deep reinforcement learning, many self-supervised learning grasping methods have been introduced.
QT-Opt, a closed-loop self-supervised reinforcement learning method for grasping using a two-fingers gripper and effective grasp cluttered objects in the real world \cite{kalashnikov2018qt}.
The works in \cite{zeng2018learning, lu2020active} used grasping and pushing in an open-loop setting using a parallel gripper.
Another work from \cite{shao2019suction} explored self-supervised learning for suction grasp in cluttered environment , but this work only considered cylinders.
While suction is often preferred over the parallel or multi-finger gripper, suction grasping draws less interest in the research community \cite{cao2021suctionnet}.
This paper aims to explore self-supervised suction grasp with cluttered objects and diverse object shapes.

\subsection{Dense Object Nets}
The ability to understand the objects with rich geometry information is critical in robotics applications.
Dense Object Nets (DONs) by \cite{florence2018dense} is a promising direction for providing such ability.
DONs is a self-supervied learning method that learns from point to point correspondence, and can provide rich information within objects.
Recent studies showed DONs were useful in robotics manipulations.
The work in  \cite{ganapathi2020learning,sundaresan2020learning} used DONs for soft material manipulation.
Another work by \cite{chai2019multi} added an additional object-centric loss to learn multi-object descriptor, which helps distinct the representation of objects better in a cluttered scene.
The descriptor was then used to solve the multi-step pick-and-place from human demonstration.
\section{Problem Definition}

In this section, we define the task, and then describe the simulation environment.

\subsection{Task Definition}

\begin{figure}[ht!b]
\vspace{2mm}
\begin{center}
  \includegraphics[width=0.9\columnwidth]{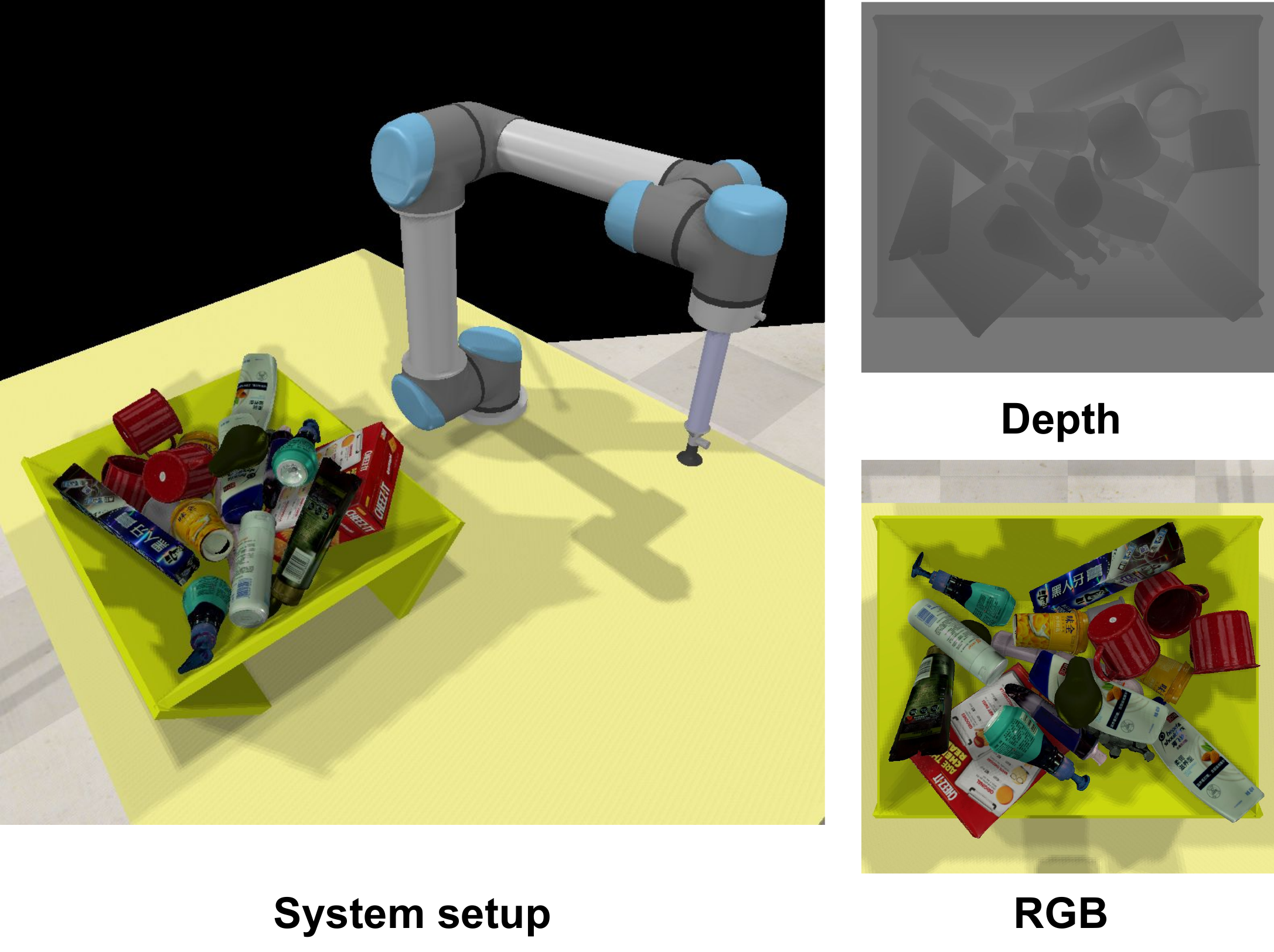}
  \caption{The simulation picking system setup, and sample RGB and depth images.}
  \label{fig:sim_env_demo}
\end{center}  
\end{figure}

We focus on picking object from a basket with cluttered general objects using a single suction pad. The input is $I = (I_{RGB}, I_D)$, where $I_{RGB}$ and $I_D$ are the RGB and depth images of the current basket. 
The action $a = (x, y)$ represent the pixel coordinate that indicates where to pick using the suction pad.
We calculate the 3D point corresponding to the selected pixel, and attempt to pick the object at the point from the surface normal direction.
At the beginning of an episode, we randomly drop objects in the basket, and we attempt to pick objects out from the basket.

\subsection{Simulation Environment}
\label{sec:sim_env}

We use CoppeliaSim \cite{coppeliaSim}, a simulation engine, and PyRep \cite{james2019pyrep}, a robotics learning toolkit, to both create the synthetic dataset for training CODs, and model the simulation environment for training and experiments.
\autoref{fig:sim_env_demo} shows our simulation environment setup. We generate dataset for training the CODs network by capturing RGB and depth images from different camera poses.
We use the UR-5 robot arm and the suction pad provided by CoppeliaSim. We modify the original suction mechanism.
See \autoref{sec:suc_mechanism} for more details.

\begin{figure*}[htb]
\vspace{1mm}
\begin{center}
  \includegraphics[width=0.75\textwidth]{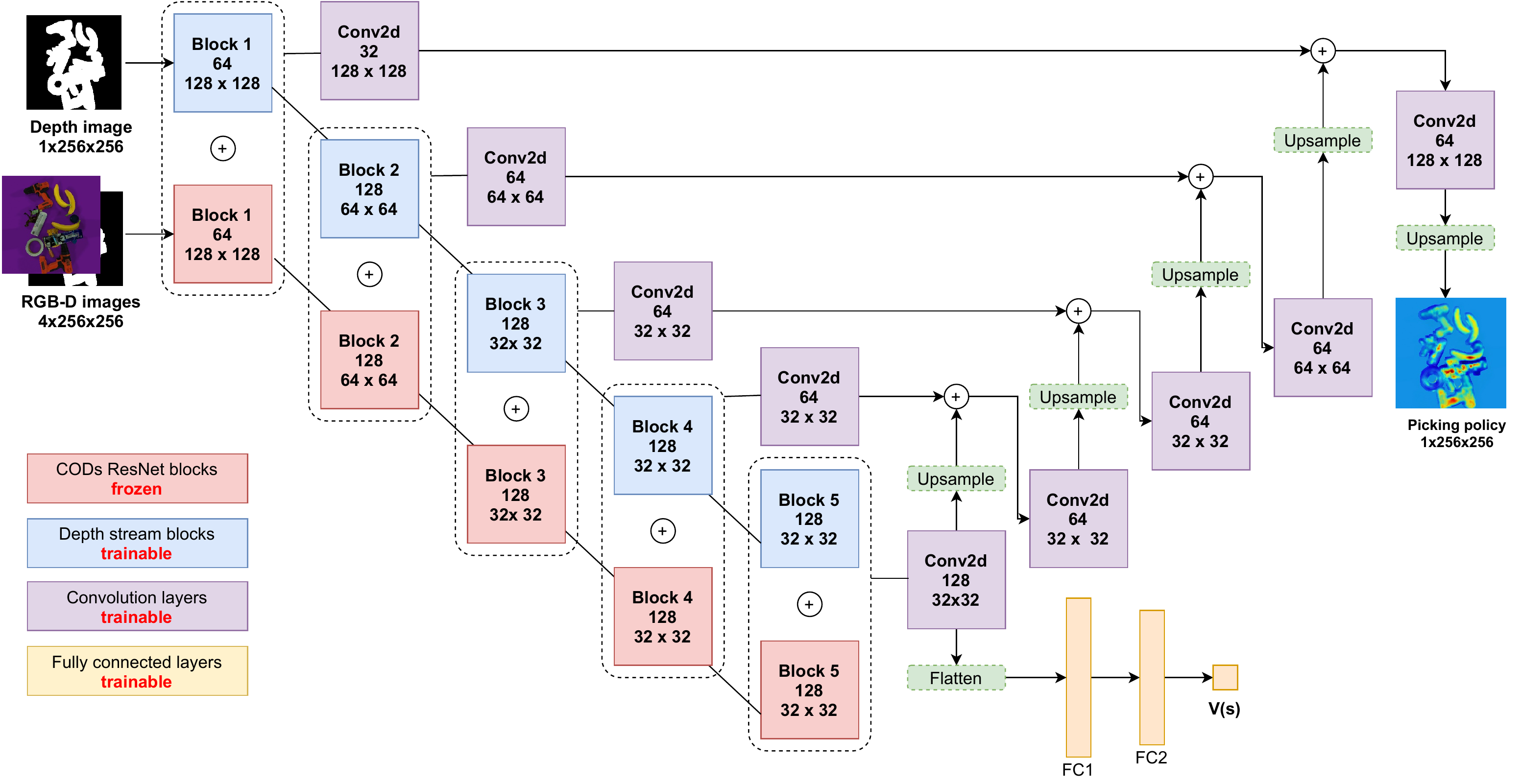}
  \caption{
  The network structure. The red boxes are the CODs stream, which are frozen during training; the blue boxes are the depth stream with trainable weights; the purple boxes are 2D convolution layers, which are connected in a U-Net fashion. There is a 2 layered multi-layer-perceptron at the bottleneck, which is the value head of the Actor-Critic method.
  }
  \label{fig:picking_network_structure}
\end{center}  
\end{figure*}

\section{Method}

In this section, we describe our method. We first present the CODs trainig, which maps the input to the dense visual descriptors for cluttered objects.
Then, we train the picking network using a reinforcement learning method. The training pipeline is shown in \autoref{fig:training_pipeline}.

\subsection{Cluttered Objects Descriptor}

Our method for training the CODs network for cluttered objects is mainly inspired by DONs \cite{florence2018dense}.
We employ the self-supervised contrastive loss from DONs on cluttered objects with randomization.

\subsubsection{Self-Supervised Contrastive Loss}

We use the contrastive loss from \cite{florence2018dense}.
Given an input image, $I \in \mathbb{R}^{W \times H \times X}$ where $X$ can be either $1, 3, 4$ depending whether the input is depth, RGB, or RGB and depth, we map $I$ to a dense descriptor space $\mathbb{R}^{W \times H \times D}$. For each pixel in the input image, we have a $D$-dimensional descriptor vector.
For a pair of inputs $I_a$ and $I_b$ captured by cameras of different poses on the same same fixed objects, we can find pairs of matching pixels, where the pixels in two images corresponds to the same vertex in the 3D reconstruction.
The dense descriptor network, $f$, is trained via a pixel wise contrastive loss to minimize the distances between descriptors of matching pixels, and keep descriptors of non-matching pixels at least $M$ distance apart. The loss function is

\begin{equation}
\mathcal{L}_{\textrm{\small{m}}}(I_a, I_b) = \frac{1}{N_{\textrm{\small{m}}}} \sum\limits_{N_{\textrm{\footnotesize{m}}}}  \lVert f(I_a)(u_a) - f(I_b)(u_b) \rVert^2_2
\end{equation}

\begin{equation}
\label{eq:non-match}
\resizebox{\hsize}{!}{
$
\mathcal{L}_{\textrm{\small{nm}}}(I_a, I_b) = \frac{1}{N_{\textrm{\small{nm}} > 0}} \sum\limits_{N_{\textrm{\footnotesize{nm}}}}  max(0, M - \lVert f(I_a)(u_a) - f(I_b)(u_b) \rVert_2)^2
$}
\end{equation}

\begin{equation}
\mathcal{L}(I_a, I_b) = \mathcal{L}_{\textrm{\small{m}}}(I_a, I_b) + \mathcal{L}_{\textrm{\small{nm}}}(I_a, I_b)
\end{equation}
where "m" represents match, and "nm" represents non-match; $N_m$ is the number of matches $(u_a, u_b)$ pixel $u_a$ in image $I_a$ and pixel $u_b$ in image $I_b$, and $N_{nm}$ is the number of non-matches $(u_a, u_b)$; $f(I)(u)$ represents the descriptor of $I$ at pixel coordinate $u \in \mathbb{N}^2$; and $N_{\textrm{nm} > 0}$ represents the number of non-zero terms in the summation term in \autoref{eq:non-match}.

\subsubsection{Data Generation and Randomization}
\label{sec:data_gen}

To generate the dataset for training CODs, we capture RGB-D images of cluttered objects in a workspace similar to our picking workspace.
We randomly sample 1 to 15 objects from the object dataset, and drop them uniformly above the workspace.
We then capture RGB-D images from different camera poses.
Additionally, we make use of randomization to help CODs learn object geometries rather than the textures, since object geometries are critical for suction grasping.
We randomize object textures and the workspace (a table) texture before capturing each RGB-D image.
For each dataset, we generate 170 static scenes, containing 20 single object scenes and 150 cluttered object scenes with 2-15 objects.
For each scene, we capture 80 RGB-D images from different camera poses.

\subsection{Picking Cluttered Objects}

\subsubsection{Picking Policy}

We use an Actor-Critic algorithm, A3C \cite{mnih2016asynchronous}, as the reinforcement learning method to train our picking policy network. Inspired by Shao \cite{shao2019suction}, we better utilize the representation power of the CODs network by using its intermediate outputs.
Shao fed the outputs of ResNet blocks to a U-Net like network. Please refer to the original paper \cite{shao2019suction} for more details.
Similarly to Shao, we feed the intermediate outputs of the CODs network, a ResNet, to a U-Net like structure, as shown in \autoref{fig:picking_network_structure}.
Additionally, we have another stream of ResNet blocks for the depth input.
As the result, we have two streams of ResNet blocks, one of which is the pre-trained CODs network.
We concatenate the corresponding outputs of ResNet blocks and, feed them forward through a U-Net like structure.
Unlike Shao, the weights of the CODs network are frozen during training.
We also add a multi-layer-perceptron at the bottleneck of the U-Net structure for value in the Actor-Critic method.
By using reinforcement learning, our method not only learns how to grasp, but also learns to avoid collisions and suction grasps that would cause invalid robot arm configurations.

\subsubsection{Picking Mechanism}
\label{sec:suc_mechanism}

\begin{SCfigure}
  \vspace{2mm}
  \caption{The modified suction pad and proximity sensors. An object is considered successfully grasped if all 7 7mm ray proximity sensors detects the object.}
  \includegraphics[width=0.4\columnwidth]{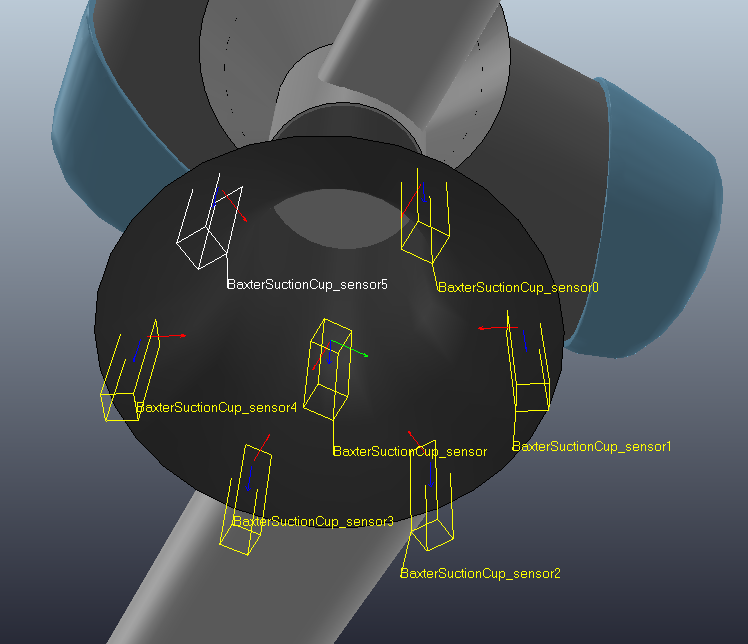}
  \label{fig:suction_cup}
\end{SCfigure}

We modify the original suction cup from CoppeliaSim to mimic the real-world suction pad more closely.
As shown in \autoref{fig:suction_cup}, in addition to a single ray proximity sensor on the center in the original suction pad, we add 6 evenly spaced ray proximity sensors near the border of the suction pad.
An object is considered successfully picked if all 7 proximity sensors detects the object. All ray proximity sensors are 7mm.

Similar to Zeng \cite{zeng2018robotic}, we consider the surface normals for picking.
After selecting an action $a=(x, y)$, a pixel coordinate in the image space for picking.
We calculate the point cloud, and estimate surface normals using Open3D \cite{zhou2018open3d}.
We obtain the 3D point $p$ and the surface normal vector $n$ corresponding to the pixel specified by $a$.
To prevent the robot arm from picking from near horizontal directions, we clip $n$ to be at most $60$ degrees from the up-right direction, and obtain the clipped vector $n'$.
The suction pad approaches $p$ from $n'$ direction, and attempt to pick an object.

\section{Experiment}

In this section, we conduct experiments to evaluate
\begin{inparaenum}[1)]
\item the effectiveness of CODs with different input configurations, and
\item the performance of the picking policy using CODs.
\end{inparaenum}

\subsection{Simulation and Dataset}

\begin{figure}[htb]
\vspace{2mm}
    \centering
    \begin{subfigure}[t]{0.32\columnwidth}
        \centering
        \includegraphics[width=\columnwidth]{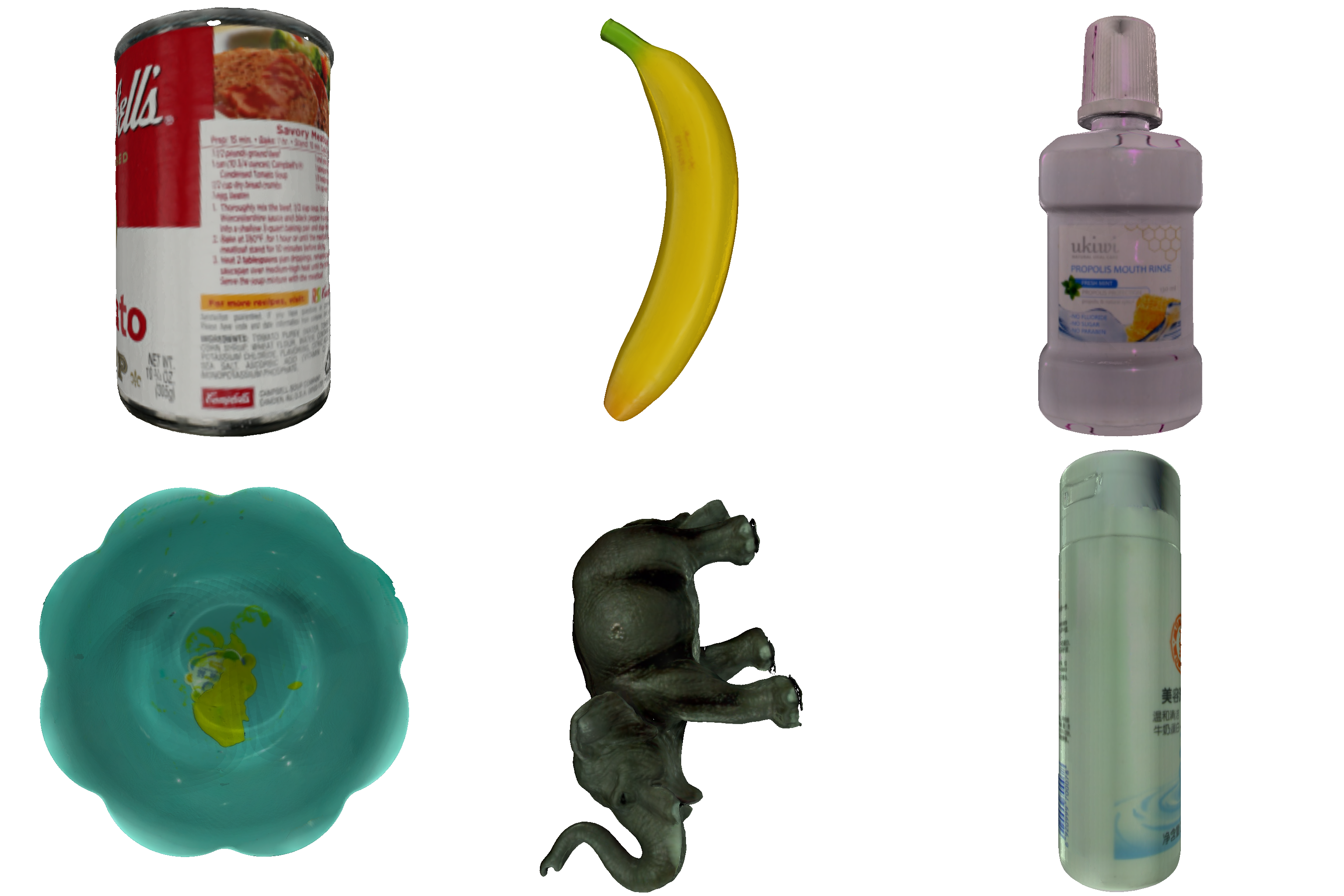}
        \caption{}
    \end{subfigure}
    \begin{subfigure}[t]{0.32\columnwidth}
        \centering
        \includegraphics[width=\columnwidth]{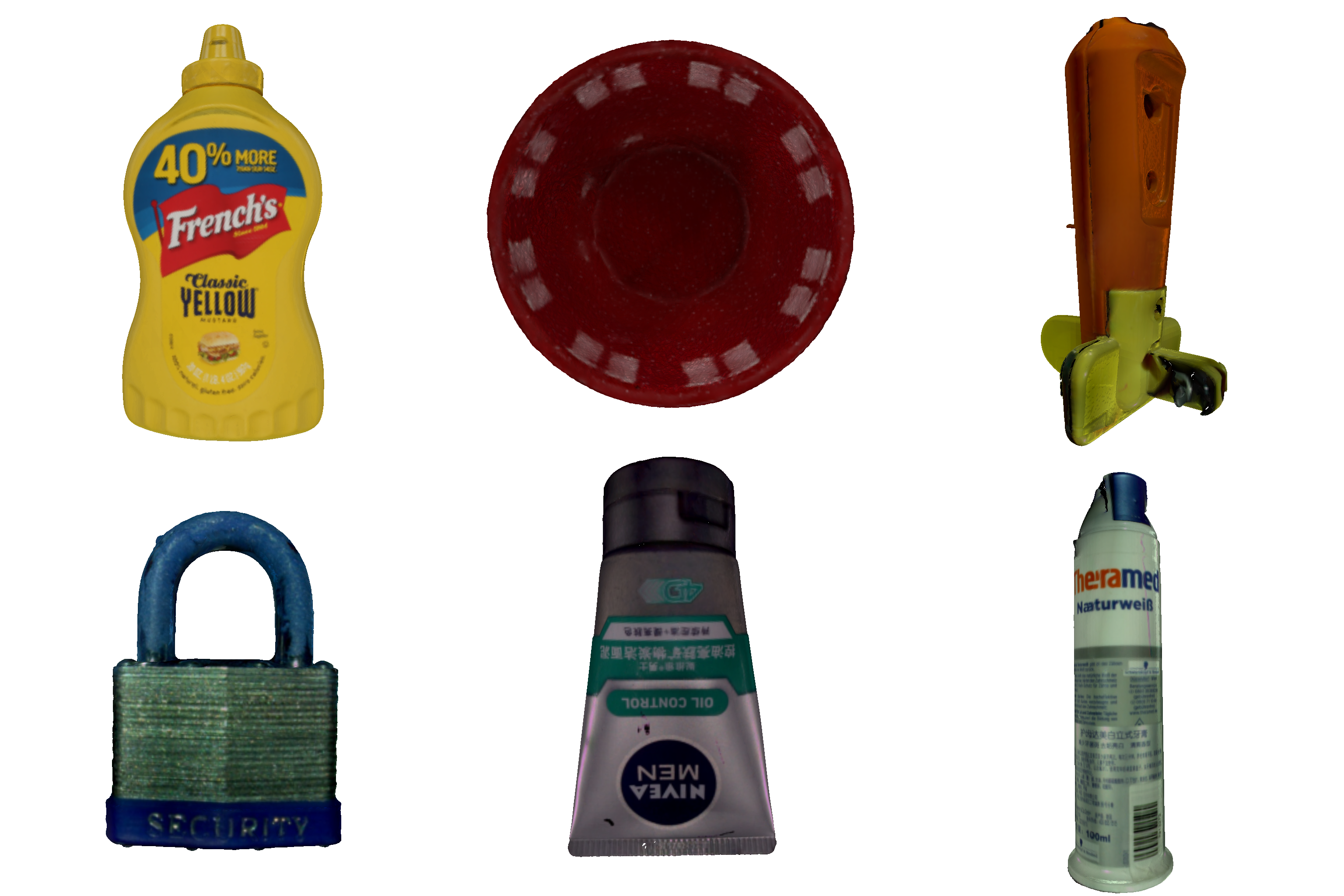}
        \caption{}
    \end{subfigure}
    \begin{subfigure}[t]{0.32\columnwidth}
        \centering
        \includegraphics[width=\columnwidth]{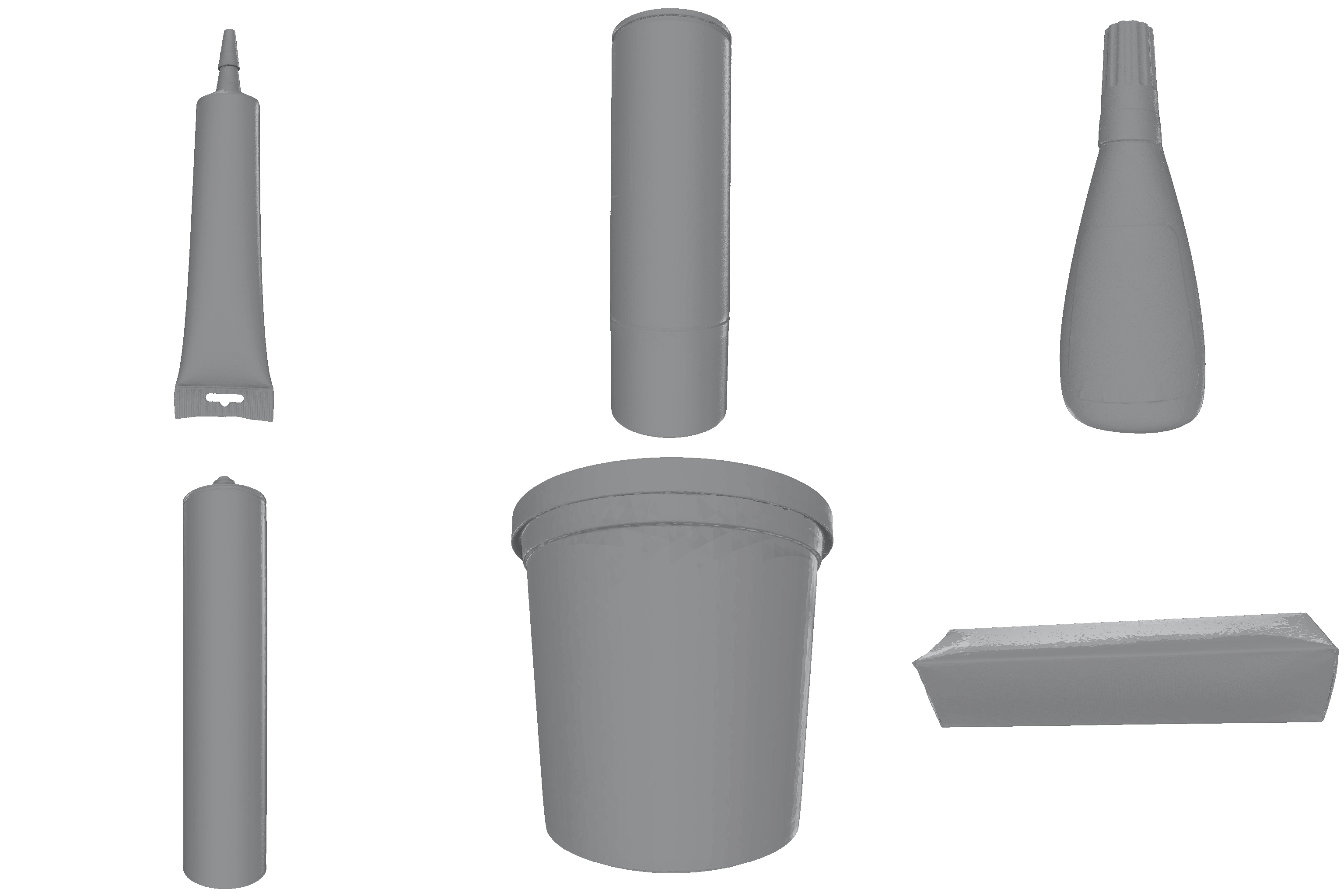}
        \caption{}
    \end{subfigure}
    \begin{subfigure}[t]{0.9\columnwidth}
        \centering
        \includegraphics[width=\columnwidth]{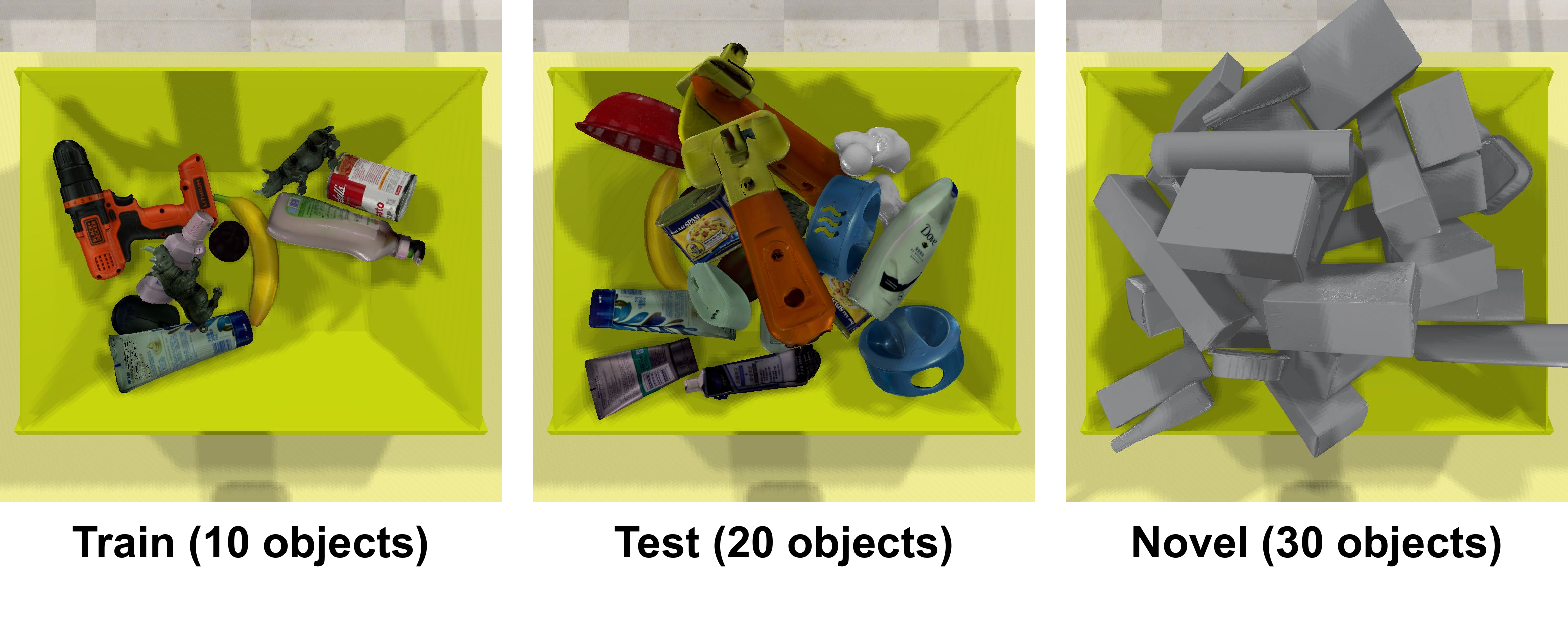}
        \caption{}
    \end{subfigure}
    \caption{Sample objects. (a) GraspNet train split. (b) GraspNet test split. (c) Novel household objects. (d) Each dataset with 10, 20, 30 objects.}
    \label{fig:dataset}
\end{figure}

\begin{itemize}

\item  \textbf{Simulation.}
We set up the simulation environment using CoppeliaSim \cite{coppeliaSim}. Please refer to \autoref{sec:sim_env} for more details. We use it for both training and testing the CODs and the picking policy.

\item  \textbf{Dataset}.
We evaluate our method on 3 datasets.
\begin{inparaenum}[1)]
\item 28 objects from the GraspNet train split, and
\item 47 objects from the GraspNet test split, and
\item 13 novel household objects \cite{fang2020graspnet}.
\end{inparaenum}
See \autoref{fig:dataset} for examples of objects from each dataset, and different levels of clutteredness. We select 55 objects that are suitable for suction grasping from the GraspNet objects.
In this paper, the GraspNet train and test splits refer to the selected objects in the original GraspNet train and test split.

\end{itemize}

\subsection{Cluttered Dense Descriptors}

\begin{table*}

\vspace{1mm}

\caption{Result of Randomization}
\label{tab:don_exp}
\centering
\begin{tabular}{|c|c|c|c|c|c|}
\hline
\multirow{2}{*}{Objects} & \multicolumn{5}{c|}{Matching Error Distance} \\
\cline{2-6}
 & RGBD w/ rand & RGB w/ rand & Depth & RGBD &  RGB \\
\hline
GraspNet (train split)    & \textbf{0.0311}  & 0.0331 & 0.0332   & 0.095 &    0.098 \\
\hline
GraspNet (test split)    & \textbf{0.0456}  & 0.0506 & 0.0518   & 0.120 &    0.121 \\
\hline
Novel       & \textbf{0.0638}  & 0.0741 & 0.0747   & 0.138 &    0.147 \\
\hline
\end{tabular}
\end{table*}

\begin{table*}
\caption{Result of Picking (Completion)}
\label{tab:picking_exp_completino}
\centering
\begin{tabular}{|c|c|c|c|c|c|c|c|c|c|c|c|c|}
\hline
{Dataset} & {\#objects} & {Shao \cite{shao2019suction} } &  {VPG Net \cite{zeng2018learning}} &  {Direct CODs} &  {Depth Only} & {CODs Only} & {CODs + Depth} \\
\hline
GraspNet  & 20    &  34.00\%   &  43.33\%  &  46.66\%  & 91.30\%  & 71.42\%   & \textbf{96.66\%}   \\
\cline{2-8}
(test split) &  30 &   23.90\%   & 23.33\%   &  34.48\%  & 78.57\% &  92.00\%  &    \textbf{93.33\%}  \\
\hline
\multirow{2}{*}{Novel objects} & 20    &  14.89\%    &  46.66\%  & 46.66\%   &  70.21\%  &  82.10\%  &    \textbf{96.69\%} \\
\cline{2-8}
 &  30 &   4.65\%   & 46.66\%   & 46.42\%   & 62.5\%   & 36.0\%  &    \textbf{68.90\%}   \\
\hline
\end{tabular}
\end{table*}

\begin{table*}
\caption{Result of Picking (Average Picked Object)}
\label{tab:picking_exp_avg_picked}
\centering
\begin{tabular}{|c|c|c|c|c|c|c|c|c|c|c|c|c|}
\hline
{Dataset} & {\#objects} & {Shao \cite{shao2019suction} } &  {VPG Net \cite{zeng2018learning}} &  {Direct CODs} &  {Depth Only} & {CODs Only} & {CODs + Depth} \\
\hline
GraspNet  & 20    &  16.12    & 15.80   &  16.45  &  19.32  &  17.75  &   \textbf{19.50}  \\
\cline{2-8}
(test split) &  30 &   22.69  & 22.23   &  23.59  &  27.50  &  28.80   &    \textbf{29.10}   \\
\hline
\multirow{2}{*}{Novel objects} & 20    &   13.11   &  17.25  &  16.90   &  18.14  & 18.30   &   \textbf{18.87}  \\
\cline{2-8}
 &  30 &   21.32   &  24.00  &  24.53  &  \textbf{27.49}  &  25.96  &    27.20  \\
\hline
\end{tabular}
\end{table*}

\begin{table*}
\caption{Result of Picking (Success Rate)}
\label{tab:picking_exp_succ_rate}
\centering
\begin{tabular}{|c|c|c|c|c|c|c|c|c|c|c|c|c|}
\hline
{Dataset} & {\#objects} & {Shao \cite{shao2019suction} } &  {VPG Net \cite{zeng2018learning}} &  {Direct CODs} &  {Depth Only} & {CODs Only} & {CODs + Depth} \\
\hline
GraspNet  & 20   &  49.10\%    &  49.07\%  &  50.95\%  & \textbf{66.20\%}  &  59.64\%  &  64.90\% \\
\cline{2-8}
(test split) &  30 &  47.70\%    &  50.05\%  &  48.74\%  &  \textbf{71.35\%}    &   57.23\%  &   63.71\%     \\
\hline
\multirow{2}{*}{Novel objects} & 20    &   31.90\%  &  47.56\%  &  47.56\%  &   62.80\% & 62.08\%    &  \textbf{66.20\%} \\
\cline{2-8}
 &  30 &  38.11\%   & 48.66\%   &  45.20\%  &  61.00\%    &  55.56\%  &   \textbf{68.20\%}    \\
\hline
\end{tabular}
\end{table*}

\subsubsection{Cluttered Objects Descriptors quality evaluation}
Similar to \cite{sundaresan2020learning}, we evaluate CODs by calculating the matching error distance normalized by image diagonal distance.
Given a pair of source and target images and a point $p$ on the source image, let $p'$ indicate the true match point in the target image, and $p^*$ indicate the best match point by using a trained descriptor.
The matching error distance is the averaged normalized distance between all pairs of $p'$ and $p^*$.
For each experiment, we sample 1000 pairs of images from the dataset, and we sample 100 pair of matching pixels on the objects from each image.


\subsubsection{Training}
\begin{figure}
    \centering
    \begin{subfigure}[t]{0.49\columnwidth}
        \includegraphics[width=\columnwidth]{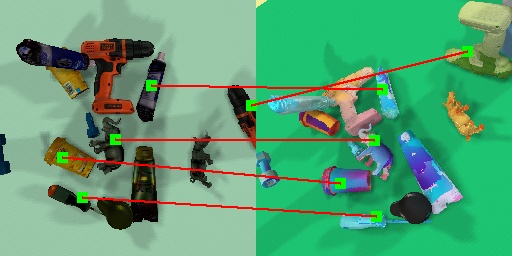}
        \caption{Match pixels.}
        \label{fig:matchPoint}
    \end{subfigure}
    \begin{subfigure}[t]{0.49\columnwidth}
        \includegraphics[width=\columnwidth]{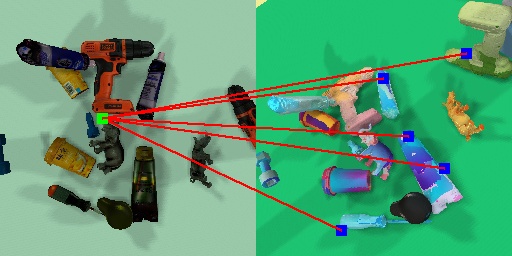}
        \caption{Object-object non-match pixels.}
        \label{fig:objectNonMatchPoint}
    \end{subfigure}
    \caption{Data generation. Match and non-match pixels with domain randomization.}
    \label{fig:dataGeneration}
\end{figure}

We use the ResNet34\_8s as the CODs network structure, same as \cite{florence2018dense}.
For each pair of images, we sample 100 pairs of match pixels on the objects, and 1500 pairs of non-match pixels for each of the three type of non-matches: object-to-object, object-to-background, background-to-background.
See \autoref{fig:dataGeneration} for examples of pairs of match and non-match pixels. See \autoref{sec:data_gen} for details for generating the datasets. The networks are trained for 120k iterations using the Adam optimizer \cite{kingma2014adam} with a weight decay of $1e^{-4}$ and a batch size of 1 on a single Nvidia GTX-1080 Ti and a Xeon CPU at 2GHz. The learning rate was set to $1e^{-1}$, and it decays by $0.9$ every 5k iterations. The descriptor vector dimension is 8, and $M$ is 0.5.

\subsubsection{Result}
We believe the depth information is important for suction grasp.
To find the best input configuration to represent cluttered objects, we compare the impacts of different input configurations: depth, RGB, RGB-D, with and without texture randomization.

\begin{figure}
    \centering
    \begin{subfigure}[t]{0.49\columnwidth}
        \includegraphics[width=\columnwidth]{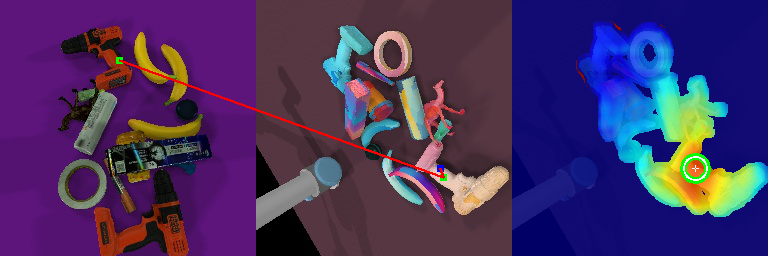}
        \caption{Matching correspondence.}
        \label{fig:matchingIndicator}
    \end{subfigure}
    \begin{subfigure}[t]{0.49\columnwidth}
        \includegraphics[width=\columnwidth]{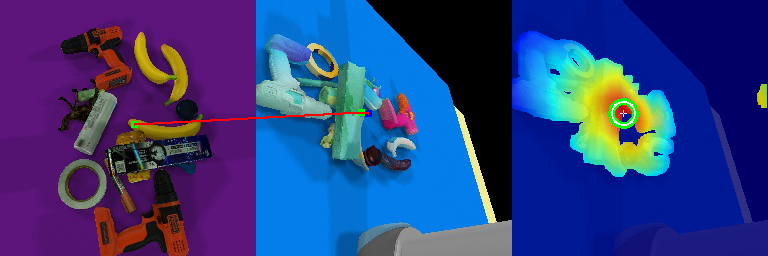}
        \caption{Matching hidden objects.}
        \label{fig:matchingIndicatorHiddenfObject}
    \end{subfigure}
    \caption{
    Correspondence evaluation. We marked the ground-truth matching pixels by green dots, connected with a line. Blue dots marked the best match pixels according to CODs. The heatmaps (in the right) indicate how much each pixel of the middle images match with the selected pixel in the left images}
    \label{fig:matchingResult}
\end{figure}

\begin{figure}[htb]
    \centering
    \begin{subfigure}[t]{0.48\columnwidth}
        \includegraphics[width=\columnwidth]{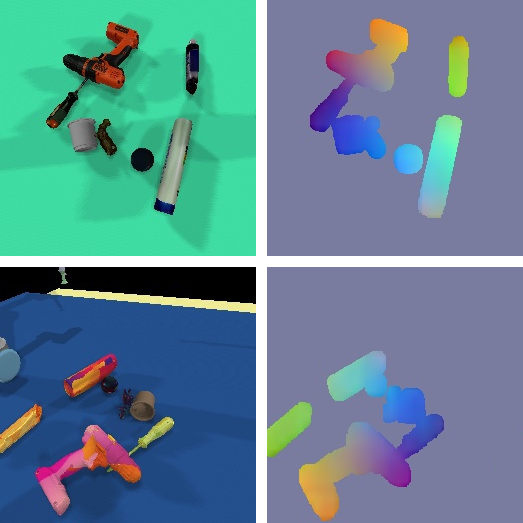}
        \caption{}
    \end{subfigure}
    \begin{subfigure}[t]{0.48\columnwidth}
        \includegraphics[width=\columnwidth]{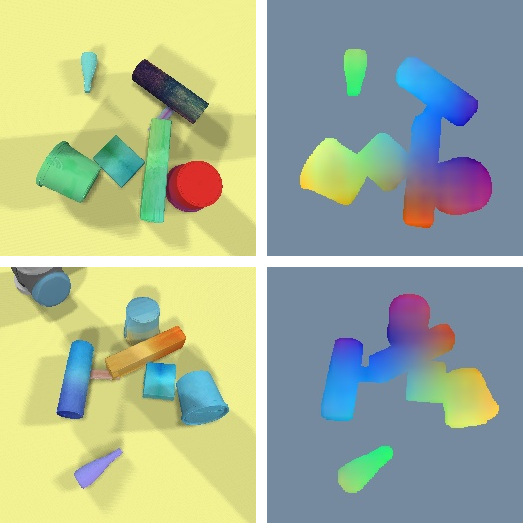}
        \caption{}
    \end{subfigure}
    \caption{
    Our CODs are consistent to represent the seen(a) and unseen(b) objects under texture randomization.
    }
    \label{fig:consitencyDescriptor}
\end{figure}

As shown in \autoref{tab:don_exp}, RGB-D with randomization outperforms other input and randomization configurations on both the GraspNet test split objects and the novel objects.
We achieve 4.56\% and 6.38\% of matching distance on GraspNet test split objects and novel objects.
The resulting CODs performs well and robust on both seen and unseen objects.
It can find matching pixels under texture randomization and occlusion, as shown in \autoref{fig:matchingResult}.
CODs can consistently represent cluttered objects with different view points on both seen and unseen objects. As shown in \autoref{fig:consitencyDescriptor}, the same parts of the same object have similar representations, invariant to viewing angles and occlusions.
Based on the experiment results, we use the CODs with RGB-D and randomization in our picking network.





\subsection{Picking Cluttered General Objects}

\subsubsection{Terminal conditions}
\label{sec:term_cond}
An episode terminates upon fulfillment of any of the following conditions.
\begin{inparaenum}[a)]
    \item{All objects have been successfully picked.}
    \item{Number of actions exceed 2 times of the number of objects at the beginning of the episode.}
    \item{The robot arm is un-controllable or unsafe to operate with. For example, the robot arm collides with the basket, the table, or itself, and the robot arm controller fails.}
\end{inparaenum}

\subsubsection{Evaluation Metrics}
We have three metrics for evaluating the performance:
\begin{inparaenum}[a)]
    \item{The \% of completion runs over all runs. A completion run is a run that all objects are picked before the episode terminates.}
    \item{The average number of objects picked in all runs.}
    \item{The average \% number of successful picks per completion run.}
\end{inparaenum}
    
\subsubsection{Baselines}

\begin{figure}[t!]
\vspace{1mm}
\begin{center}
  \includegraphics[width=\columnwidth]{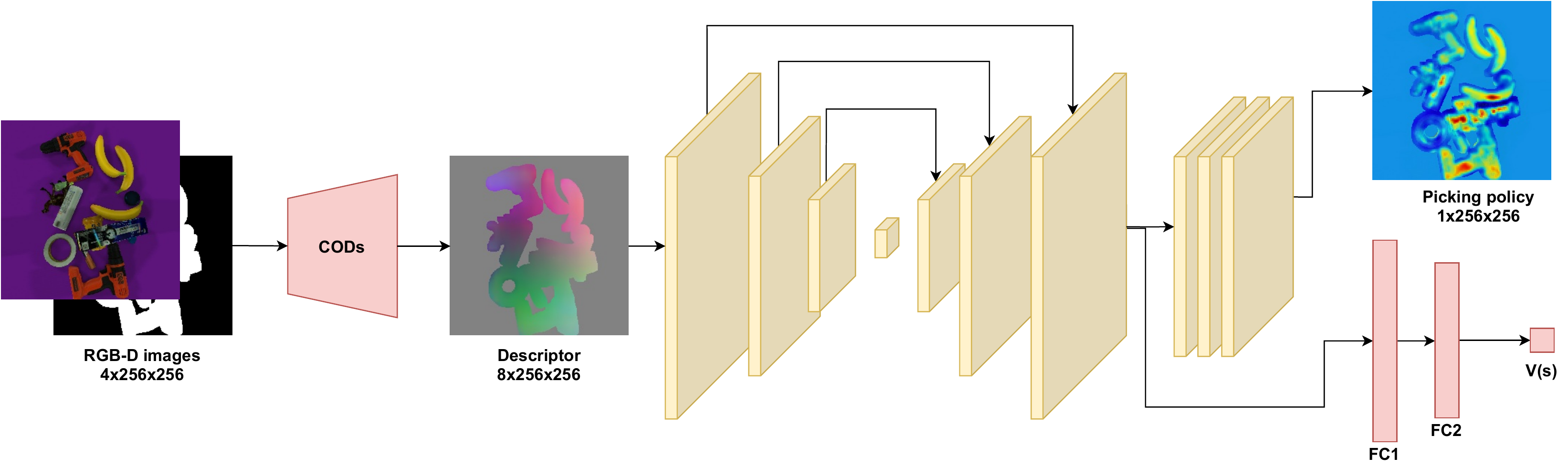}
  \caption{The network structure that directly feeds output of the CODs network to a U-Net.}
  \label{fig:picking_network_structure_old}
\end{center}  
\end{figure}

We compare with the following methods:
\begin{itemize}
    \item Suction Grasp Region Prediction: Shao \cite{shao2019suction} proposed a network structure that combines ResNet34-stride8 and U-Net.
    We use Shao's structure in our training pipeline with the same environment and hyper-parameters.
    
    \item VPG Net: Zeng's \cite{zeng2018learning} fully convolutional network with DenseNet121 backbone.
    In addition, we add upsample convolutional layers to increase the output size from 8x8 to 128x128, because 8x8 was not suitable for picking cluttered objects.
    We also add two fully connected layers at the bottle neck of the network to output the value head of the Actor-Critic method.
    
    \item Direct CODs: A network structure that directly uses the output from CODs with a U-Net.
    Please refer to \autoref{fig:picking_network_structure_old} for the network structure.
    \item Depth Only: Our method using only the depth input.
    \item CODs Only: Our method using only the CODs input.
\end{itemize}
    
\subsubsection{Rewards and training}

We implement Actor-Critic with Python 3.8 and PyTorch. We use the Adam optimizer \cite{kingma2014adam} with a learning rate of 0.0005 and a momentum of 0.9. The hyper-parameters for Actor-Critic are the following: entropy coefficient is beta is 0.001, the clipping parameter epsilon is 0.2, and the discount factor is 0.3.

The rewards are as follows: $+0.1$ for each successful pick, $-0.1$ for a failed pick, and $-1$ for terminal steps. Please refer to \autoref{sec:term_cond} for details about the terminal conditions.
We use 4 parallel simulation environments to train the policies on a single Nvidia GTX-1080 Ti and a Xeon CPU at 2GHz.

\subsubsection{Simulation Experiments}

We train each method in the simulation environment with the GraspNet train split on 10 objects.
We run 50 episodes to test each method with the GraspNet test split and novel objects on 20 and 30 objects.
Testing scenarios are much more cluttered than the training scenarios.
Please refer to \autoref{fig:dataset} for examples of objects from each dataset and different levels of clutterness.
\subsubsection{Results}

\begin{figure}[t!]
\begin{center}
  \includegraphics[width=\columnwidth]{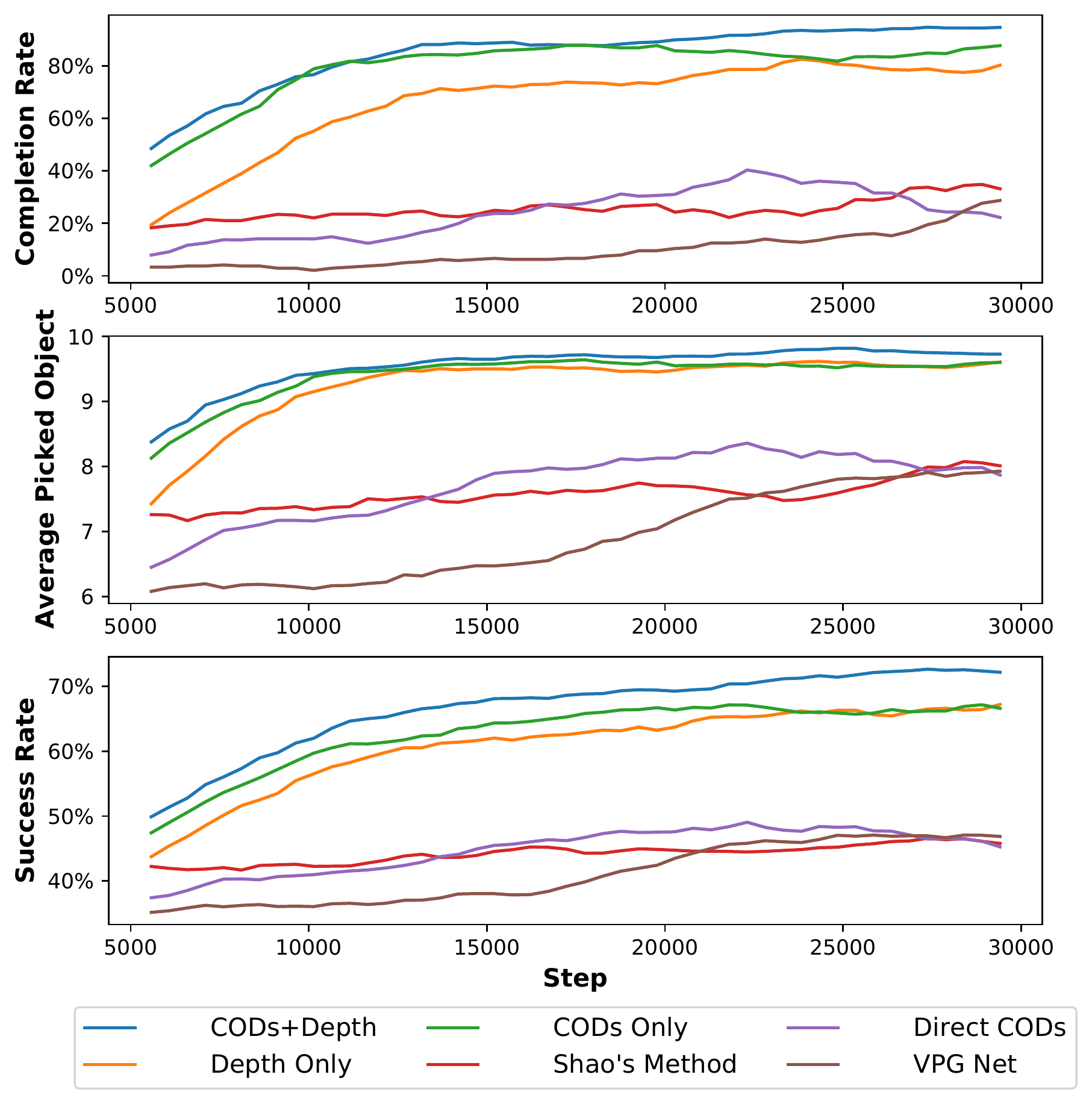}
  \caption{Evaluation metrics for all methods during training.}
  \label{fig:exp_train_plot}
\end{center}  
\end{figure}

As shown in \autoref{fig:exp_train_plot}, our method can learn picking faster and better than other methods during training.
As shown in \autoref{tab:picking_exp_completino}, \autoref{tab:picking_exp_avg_picked}, and \autoref{tab:picking_exp_succ_rate}, our method outperforms other methods on most of the metrics.
Especially in the case of picking unseen objects which is twice as cluttered as the training scenarios, our method reaches 96.69\% completion.
The model that uses both the depth stream and CODs stream, performs better than other methods that uses only the depth stream and Shao's method by a large margin.
In addition, our picking policy network which uses the intermediate outputs of the CODs network, out-matches the method that only uses the final output.

\section{Conclusion}

In this paper, we extend the Dense Object Descriptors \cite{florence2018dense} to better represent cluttered objects, called the CODs.
CODs is able to robustly represent cluttered objects under texture and viewing angles changes.
Hence, CODs can focus on the object geometries, which is a useful characteristic for suction grasping.
We utilize the intermediate outputs of the CODs network to improve the picking performance.
As the result, our method can learn to effectively pick cluttered general objects while avoiding collisions and control failures.
The resulting policy can pick 96.69\% of unseen objects that  are  twice  as  cluttered  as  the  training  scenarios.
In the future works, we will extend applications of CODs to more robotics manipulations tasks where the geometry information is critical.






\section*{ACKNOWLEDGMENT}
This research is partially supported by the Ministry of Science and Technology (MOST) of Taiwan under Grant Numbers 110-2634-F-009-022, 110-2634-F-A49-004 and 110-2221-E-A49-067-MY3, and the computing resources are partially supported by National Center for High-performance Computing (NCHC) of Taiwan. 


\bibliographystyle{plain}
\bibliography{egbib}

\begin{thebibliography}{10}

\bibitem{cao2021suctionnet}
Hanwen Cao, Hao-Shu Fang, Wenhai Liu, and Cewu Lu.
\newblock Suctionnet-1billion: A large-scale benchmark for suction grasping.
\newblock {\em arXiv preprint arXiv:2103.12311}, 2021.

\bibitem{chai2019multi}
Chun-Yu Chai, Keng-Fu Hsu, and Shiao-Li Tsao.
\newblock Multi-step pick-and-place tasks using object-centric dense
  correspondences.
\newblock In {\em 2019 IEEE/RSJ International Conference on Intelligent Robots
  and Systems (IROS)}, pages 4004--4011. IEEE, 2019.

\bibitem{danielczuk2020xray}
Michael Danielczuk, Anelia Angelova, Vincent Vanhoucke, and Ken Goldberg.
\newblock X-ray: Mechanical search for an occluded object by minimizing support
  of learned occupancy distributions, 2020.

\bibitem{du2021vision}
Guoguang Du, Kai Wang, Shiguo Lian, and Kaiyong Zhao.
\newblock Vision-based robotic grasping from object localization, object pose
  estimation to grasp estimation for parallel grippers: a review.
\newblock {\em Artificial Intelligence Review}, 54(3):1677--1734, 2021.

\bibitem{du2019good}
Simon~S Du, Sham~M Kakade, Ruosong Wang, and Lin~F Yang.
\newblock Is a good representation sufficient for sample efficient
  reinforcement learning?
\newblock {\em arXiv preprint arXiv:1910.03016}, 2019.

\bibitem{fang2020graspnet}
Hao-Shu Fang, Chenxi Wang, Minghao Gou, and Cewu Lu.
\newblock Graspnet-1billion: A large-scale benchmark for general object
  grasping.
\newblock In {\em Proceedings of the IEEE/CVF Conference on Computer Vision and
  Pattern Recognition}, pages 11444--11453, 2020.

\bibitem{florence2018dense}
Peter~R Florence, Lucas Manuelli, and Russ Tedrake.
\newblock Dense object nets: Learning dense visual object descriptors by and
  for robotic manipulation.
\newblock {\em arXiv preprint arXiv:1806.08756}, 2018.

\bibitem{ganapathi2020learning}
Aditya Ganapathi, Priya Sundaresan, Brijen Thananjeyan, Ashwin Balakrishna,
  Daniel Seita, Jennifer Grannen, Minho Hwang, Ryan Hoque, Joseph~E Gonzalez,
  Nawid Jamali, et~al.
\newblock Learning dense visual correspondences in simulation to smooth and
  fold real fabrics.
\newblock {\em arXiv preprint arXiv:2003.12698}, 2020.

\bibitem{hasegawa2019graspfusion}
Shun Hasegawa, Kentaro Wada, Shingo Kitagawa, Yuto Uchimi, Kei Okada, and
  Masayuki Inaba.
\newblock Graspfusion: Realizing complex motion by learning and fusing grasp
  modalities with instance segmentation.
\newblock In {\em 2019 International Conference on Robotics and Automation
  (ICRA)}, pages 7235--7241. IEEE, 2019.

\bibitem{james2019pyrep}
Stephen James, Marc Freese, and Andrew~J Davison.
\newblock Pyrep: Bringing v-rep to deep robot learning.
\newblock {\em arXiv preprint arXiv:1906.11176}, 2019.

\bibitem{jiang2020depth}
Ping Jiang, Yoshiyuki Ishihara, Nobukatsu Sugiyama, Junji Oaki, Seiji Tokura,
  Atsushi Sugahara, and Akihito Ogawa.
\newblock Depth image--based deep learning of grasp planning for textureless
  planar-faced objects in vision-guided robotic bin-picking.
\newblock {\em Sensors}, 20(3):706, 2020.

\bibitem{kalashnikov2018qt}
Dmitry Kalashnikov, Alex Irpan, Peter Pastor, Julian Ibarz, Alexander Herzog,
  Eric Jang, Deirdre Quillen, Ethan Holly, Mrinal Kalakrishnan, Vincent
  Vanhoucke, et~al.
\newblock Qt-opt: Scalable deep reinforcement learning for vision-based robotic
  manipulation.
\newblock {\em arXiv preprint arXiv:1806.10293}, 2018.

\bibitem{kingma2014adam}
Diederik~P Kingma and Jimmy Ba.
\newblock Adam: A method for stochastic optimization.
\newblock {\em arXiv preprint arXiv:1412.6980}, 2014.

\bibitem{kumra2017robotic}
Sulabh Kumra and Christopher Kanan.
\newblock Robotic grasp detection using deep convolutional neural networks.
\newblock In {\em 2017 IEEE/RSJ International Conference on Intelligent Robots
  and Systems (IROS)}, pages 769--776. IEEE, 2017.

\bibitem{lu2020active}
Ning Lu, Tao Lu, Yinghao Cai, and Shuo Wang.
\newblock Active pushing for better grasping in dense clutter with deep
  reinforcement learning.
\newblock In {\em 2020 Chinese Automation Congress (CAC)}, pages 1657--1663.
  IEEE, 2020.

\bibitem{mahler2018dex}
Jeffrey Mahler, Matthew Matl, Xinyu Liu, Albert Li, David Gealy, and Ken
  Goldberg.
\newblock Dex-net 3.0: Computing robust vacuum suction grasp targets in point
  clouds using a new analytic model and deep learning.
\newblock In {\em 2018 IEEE International Conference on robotics and automation
  (ICRA)}, pages 5620--5627. IEEE, 2018.

\bibitem{mahler2019learning}
Jeffrey Mahler, Matthew Matl, Vishal Satish, Michael Danielczuk, Bill DeRose,
  Stephen McKinley, and Ken Goldberg.
\newblock Learning ambidextrous robot grasping policies.
\newblock {\em Science Robotics}, 4(26):eaau4984, 2019.

\bibitem{mnih2016asynchronous}
Volodymyr Mnih, Adri\`{a}~Puigdom\`{e}nech Badia, Mehdi Mirza, Alex Graves, Tim
  Harley, Timothy~P. Lillicrap, David Silver, and Koray Kavukcuoglu.
\newblock Asynchronous methods for deep reinforcement learning.
\newblock In {\em Proceedings of the 33rd International Conference on
  International Conference on Machine Learning - Volume 48}, ICML'16, page
  1928–1937. JMLR.org, 2016.

\bibitem{qin2020s4g}
Yuzhe Qin, Rui Chen, Hao Zhu, Meng Song, Jing Xu, and Hao Su.
\newblock S4g: Amodal single-view single-shot se (3) grasp detection in
  cluttered scenes.
\newblock In {\em Conference on robot learning}, pages 53--65. PMLR, 2020.

\bibitem{coppeliaSim}
E.~Rohmer, S.~P.~N. Singh, and M.~Freese.
\newblock Coppeliasim (formerly v-rep): a versatile and scalable robot
  simulation framework.
\newblock In {\em Proc. of The International Conference on Intelligent Robots
  and Systems (IROS)}, 2013.
\newblock www.coppeliarobotics.com.

\bibitem{shao2019suction}
Quanquan Shao, Jie Hu, Weiming Wang, Yi~Fang, Wenhai Liu, Jin Qi, and Jin Ma.
\newblock Suction grasp region prediction using self-supervised learning for
  object picking in dense clutter.
\newblock In {\em 2019 IEEE 5th International Conference on Mechatronics System
  and Robots (ICMSR)}, pages 7--12. IEEE, 2019.

\bibitem{sundaresan2020learning}
Priya Sundaresan, Jennifer Grannen, Brijen Thananjeyan, Ashwin Balakrishna,
  Michael Laskey, Kevin Stone, Joseph~E Gonzalez, and Ken Goldberg.
\newblock Learning rope manipulation policies using dense object descriptors
  trained on synthetic depth data.
\newblock In {\em 2020 IEEE International Conference on Robotics and Automation
  (ICRA)}, pages 9411--9418. IEEE, 2020.

\bibitem{utomo2021suction}
Tri~Wahyu Utomo, Adha~Imam Cahyadi, and Igi Ardiyanto.
\newblock Suction-based grasp point estimation in cluttered environment for
  robotic manipulator using deep learning-based affordance map.
\newblock {\em International Journal of Automation and Computing},
  18(2):277--287, 2021.

\bibitem{valencia20173d}
Angel~J Valencia, Roger~M Idrovo, Angel~D Sappa, Douglas~Plaza Guingla, and
  Daniel Ochoa.
\newblock A 3d vision based approach for optimal grasp of vacuum grippers.
\newblock In {\em 2017 IEEE International Workshop of electronics, control,
  measurement, signals and their application to mechatronics (ECMSM)}, pages
  1--6. IEEE, 2017.

\bibitem{wan2020planning}
Weiwei Wan, Kensuke Harada, and Fumio Kanehiro.
\newblock Planning grasps with suction cups and parallel grippers using
  superimposed segmentation of object meshes.
\newblock {\em IEEE Transactions on Robotics}, 37(1):166--184, 2020.

\bibitem{zeng2018learning}
Andy Zeng, Shuran Song, Stefan Welker, Johnny Lee, Alberto Rodriguez, and
  Thomas Funkhouser.
\newblock Learning synergies between pushing and grasping with self-supervised
  deep reinforcement learning.
\newblock In {\em 2018 IEEE/RSJ International Conference on Intelligent Robots
  and Systems (IROS)}, pages 4238--4245. IEEE, 2018.

\bibitem{zeng2018robotic}
Andy Zeng, Shuran Song, Kuan-Ting Yu, Elliott Donlon, Francois~R Hogan, Maria
  Bauza, Daolin Ma, Orion Taylor, Melody Liu, Eudald Romo, et~al.
\newblock Robotic pick-and-place of novel objects in clutter with
  multi-affordance grasping and cross-domain image matching.
\newblock In {\em 2018 IEEE international conference on robotics and automation
  (ICRA)}, pages 3750--3757. IEEE, 2018.

\bibitem{zhou2018open3d}
Qian-Yi Zhou, Jaesik Park, and Vladlen Koltun.
\newblock Open3d: A modern library for 3d data processing.
\newblock {\em arXiv preprint arXiv:1801.09847}, 2018.

\end{thebibliography}

\end{document}